\pdfoutput=1

\documentclass[11pt]{article}

\usepackage[final]{acl}

\usepackage{times}
\usepackage{latexsym}

\usepackage{soul}

\def\code#1{\sethlcolor{lightgray}\hl{\texttt{#1}}}

\usepackage[T1]{fontenc}

\usepackage[utf8]{inputenc}

\usepackage{microtype}

\usepackage{inconsolata}

\usepackage{graphicx}

\usepackage{amsmath}
\usepackage{amsfonts}  
\usepackage{amssymb}    

\usepackage[most]{tcolorbox}

\usepackage{xcolor}
\definecolor{lightgray}{gray}{0.92}
\definecolor{bulgarianrose}{rgb}{0.28, 0.02, 0.03}
\definecolor{PastelGold}{RGB}{255, 245, 200}
\definecolor{PastelSilver}{RGB}{220, 220, 220}
\definecolor{PastelBronze}{RGB}{240, 210, 180}
\definecolor{PastelRed}{RGB}{255, 182, 193}
\definecolor{VibrantGold}{RGB}{255, 215, 0}
\definecolor{VibrantSilver}{RGB}{172, 172, 172}
\definecolor{VibrantBronze}{RGB}{205, 127, 50}
\usepackage{fontawesome5}

\usepackage[utf8]{inputenc}
\newcommand{\block}[1]{%
	\raisebox{\dimexpr(\fontcharht\font`X-1em)/2}{\rule{1em}{#1\dimexpr1em/8}}%
	}
	\DeclareUnicodeCharacter{2581}{\block{.5}}

	\usepackage{multirow} 
	\usepackage{tabularx}
	\usepackage{makecell} 
	\usepackage{booktabs}
	\usepackage{siunitx} 
	\usepackage{threeparttable} 
	\usepackage{longtable} 

	\usepackage{caption}

	\usepackage[printonlyused,withpage,nolist]{acronym}

	\usepackage[titletoc]{appendix} 
	\usepackage{float} 
	\usepackage{placeins} 

	\title{TransBERT: A Framework for Synthetic Translation in Domain-Specific Language Modeling}

	\author{
		Julien Knafou, Luc Mottin, Ana\"{i}s Mottaz, Alexandre Flament, Patrick Ruch \\
		HES-SO, Geneva, Switzerland \\ SIB, Swiss Institute of Bioinformatics, Geneva, Switzerland
		}

\begin{document}
		\begin{acronym}
    \acro{NLP}{Natural Language Processing}
    \acro{NER}{Named Entity Recognition}
    \acro{RE}{Relation Extraction}
    \acro{MT}{Machine Translation}
    \acro{AI}{Artificial Intelligence}
    \acro{LM}{Language Model}
    \acro{PLM}{Pre-trained Language Model}
    \acro{LLM}{Large Language Model}
    \acro{SMT}{Statistical Machine Translation}
    \acro{NMT}{Neural Machine Translation}
    \acro{NN}{Neural Network}
    \acro{DS}{Domain-Specific}
    \acro{IR}{Information Retrieval}
    \acro{POS}{Part-Of-Speech}
    \acro{SRL}{Semantic Role Labeling}
    \acro{SRW}{Semantically Related Words}
    \acro{RNN}{Recurrent Neural Network}
    \acro{CNN}{Convolutional Neural Network}
    \acro{BERT}{Bidirectional Encoder Representations from Transformers}
    \acro{BLEU}{Bi-Lingual Evaluation Understudy}
    \acro{GloVe}{Global Vectors for Word Representation}
    \acro{CBOW}{Continuous Bag-of-Words}
    \acro{ELMo}{Embeddings from Language Models}
    \acro{LSTM}{Long Short Term Memory}
    \acro{QA}{Question Answering}
    \acro{SOTA}{State-of-the-Art}
    \acro{Seq2Seq}{Sequence-to-Sequence}
    \acro{MLM}{Masked Language Model}
    \acro{NSP}{Next Sentence Prediction}
    \acro{OOV}{Out-of-Vocabulary}
    \acro{GLUE}{General Language Understanding Evaluation}
    \acro{SQuAD}{Stanford Question Answering Dataset}
    \acro{SWAG}{Situations with Adversarial Generations}
    \acro{CORD-19}{COVID-19 Open Research Dataset}
    \acro{COAP}{COVID-19 Open Access Project}
    \acro{TF-IDF}{Term Frequency-Inverse Document Frequency}
    \acro{W2V}{Word2Vec}
    \acro{M2M}{Many-to-Many}
    \acro{BLURB}{Biomedical Language Understanding \& Reasoning Benchmark}
    \acro{NLU}{Natural Language Understanding}
    \acro{BLUE}{Biomedical Language Understanding Evaluation}
    \acro{BPE}{Byte Pair Encoding}
    \acro{GRU}{Gated Recurrent Units}
    \acro{RoBERTa}{Robustly optimized BERT approach}
    \acro{PMC}{PubMed Central}
    \acro{OSCAR}{Open Super-large Crawled Aggregated coRpus}
    \acro{WWM}{Whole-Word Masking}
    \acro{MCQA}{Multiple-Choice Question Answering}
    \acro{DL}{Deep Learning}
    \acro{NLG}{Natural Language Generation}
    \acro{RBMT}{Rule-Based Machine Translation}
    \acro{EBMT}{Example-Based Machine Translation}
    \acro{WMT}{Workshop on Machine Translation}
    \acro{BT}{Backtranslation}
    \acro{W-NUT}{Workshop on Noisy User-generated Text}
    \acro{DEFT}{DÉfi Fouille de Textes}
    \acro{ChEMU}{Cheminformatics Elsevier Melbourne University}
    \acro{CLEF}{Conference and Labs of the Evaluation Forum}
    \acro{UniProtKB}{UniProt Knowledgebase}
    \acro{TREC}{Text REtrieval Conference}
    \acro{EHRs}{Electronic Health Records}
    \acro{GPU}{Graphics Processing Unit}
    \acro{LLaMA}{Large Language Model Meta AI}
    \acro{GPT}{Generative Pre-trained Transformer}
    \acro{T5}{Text-to-Text Transfer Transformer}
    \acro{NIH}{National Institutes of Health}
    \acro{NLM}{National Library of Medicine}
    \acro{MBR}{MEDLINE/PubMed Baseline Repository}
    \acro{MeSH}{Medical Subject Headings}
    \acro{NCBI}{National Center for Biotechnology Information}
    \acro{SMBO}{Sequential Model-Based Optimization}
    \acro{HPO}{Hyperparameter Optimization}
    \acro{IOB}{Inside-Outside-Beginning}
    \acro{NLTK}{Natural Language Toolkit}
    \acro{STS}{Semantic Textual Similarity}
    \acro{EMR}{Exact Match Ratio}
    \acro{RMSE}{Root Mean Squared Error}
    \acro{HPC}{High-Performance Computing}
    \acro{ANOVA}{Analysis Of Variance}
    \acro{CI}{Confidence Interval}
    \acro{AUC}{Area Under the Cruve}
    \acro{NRA}{Normalized Ranking Average}
    \acro{PPPL}{Pseudo-Perplexity}
    \acro{PLLs}{Pseudo-Log-Likelihood scores}
    \acro{ALM}{Autoregressive Language Model}
    \acro{CmBERT}{CamemBERT}
	\acro{CLI}{Command Line Interface}
	\acro{API}{Application Programming Interface}
	\acro{TransBERT}{TransBERT-bio-fr}
\end{acronym}

		\maketitle
		\begin{abstract}
			The scarcity of non-English language data in specialized domains significantly limits the development of effective Natural Language Processing (NLP) tools. We present TransBERT, a novel framework for pre-training language models using exclusively synthetically translated text, and introduce TransCorpus, a scalable translation toolkit. Focusing on the life sciences domain in French, our approach demonstrates that state-of-the-art performance on various downstream tasks can be achieved solely by leveraging synthetically translated data. We release the TransCorpus toolkit, the TransCorpus-bio-fr corpus (36.4GB of French life sciences text), TransBERT-bio-fr, its associated pre-trained language model and reproducible code for both pre-training and fine-tuning. Our results highlight the viability of synthetic translation in a high-resource translation direction for building high-quality NLP resources in low-resource language/domain pairs.
		\end{abstract}

		\section{Introduction}
		\acp{PLM} have revolutionized the field of \ac{NLP} by leveraging large-scale datasets and powerful neural network architectures to learn rich linguistic representations. These models, such as BERT \cite{devlin-etal-2019-bert}, GPT \cite{radford2018improving}, and T5 \cite{DBLP:journals/corr/abs-1910-10683}, are pre-trained on vast amounts of text data in an unsupervised manner, enabling them to capture intricate patterns and nuances of human language. \acp{PLM} can be fine-tuned for specific tasks, such as text classification, \ac{NER}, and \ac{QA}, by training them on smaller labeled datasets. This transfer learning approach has significantly improved the performance of \ac{NLP} models across various languages and domains.

		Unfortunately, the success of \acp{PLM} has not been equally distributed across all languages. While high-resource languages like English, Chinese, and French have seen significant advancements in \ac{NLP} applications, many low-resource languages still lack the necessary data and resources to develop effective models. This disparity is particularly evident in specialized domains such as life sciences, where the availability of high-quality datasets is crucial for training accurate models. For example, Hindi, which is spoken by over 600M people, has no available \ac{PLM} for the life sciences domain. Although BioBERT \cite{Lee_2019}, the first pre-trained language model for the life sciences, was released in 2019, significant efforts to gather sufficient \ac{DS} data for training life sciences models in other high-resource languages have begun to emerge in recent years. Since 2023, life sciences models have emerged for German \cite{de_biobert}, Italian \cite{it_biobert}, and French \cite{labrak-etal-2023-drbert,touchent2023camembertbio}. Life sciences is only an example of a domain where the lack of data is a significant barrier to the development of \ac{NLP} tools. Other domains, such as legal, finance, and patent, also face similar challenges.

		In this paper, we introduce TransCorpus, an open-source toolkit leveraging the fairseq translation framework \cite{fairseq} to generate extensive synthetic \ac{DS} corpora in up to 100 languages, featuring a production-level API/CLI setup with multi-GPU and multi-processing capabilities, and reliable checkpoint recovery for scalable corpus management. In the context of a high-resource translation direction, we demonstrate that a \ac{LM} trained on TransCorpus output can achieve state-of-the-art performance on various downstream tasks by leveraging DrBenchmark, a French life sciences benchmark \cite{labrak:hal-04470938}.

		Our contributions are threefold with (1) the release of an open-source toolkit\footnote{\url{https://github.com/jknafou/TransCorpus}} for scalable multilingual corpus translation, (2) a French life sciences corpus \cite{knafou2026transcorpusbio}\footnote{\url{https://huggingface.co/datasets/jknafou/TransCorpus-bio}} synthetically translated of 36.4GB along with a tokenizer, and a \ac{PLM} \cite{knafou2026transbertbiofr}\footnote{\url{https://huggingface.co/jknafou/TransBERT-bio-fr}} released on Hugging Face, and (3) reproducible code for the pre-training and fine-tuning of the French life sciences \ac{PLM} made publicly available on GitHub\footnote{\url{https://github.com/jknafou/TransBERT}}. For future research, we are in the process of adding new languages and publishing them in the Hugging Face datasets repository. To see the current state of the added languages, check out the following link: \url{https://huggingface.co/datasets/jknafou/TransCorpus-bio}.

		\section{Related Work}
		The paradigm of training \acp{LM} on massive datasets is relatively recent, gaining prominence after the introduction of BERT \cite{devlin2018bert}, and as a result, there is still limited research leveraging translation for training such models, especially in low-resource settings.

		In \citet{isbister2021stop}, sentiment analysis in four low-resource Scandinavian languages is explored using three strategies: fine-tuning a native monolingual \ac{PLM}, translating the data into English and fine-tuning an English \ac{PLM}, and fine-tuning a multilingual \ac{PLM} on the native data. Results generally favor the multilingual approach, though fine-tuning an English model on translated data often outperforms using a monolingual low-resource \ac{PLM}.

		For Luxembourgish, \citet{lothritz-etal-2022-luxembert} tackle data scarcity by partially translating unambiguous words from a related high-resource language, evaluating several models including a Luxembourgish-only BERT, a Luxembourgish-German BERT, and LuxemBERT, which is trained on mixed corpora. LuxemBERT shows improved performance over mBERT, though not to a statistically significant extent.

		In the Basque context, following the introduction of ElhBERTeu~\cite{urbizu-etal-2022-basqueglue}, \citet{urbizu-etal-2023-enough} use synthetic translated data from Spanish to enlarge the Basque corpus, finding that while a \ac{PLM} trained solely on synthetic data is competitive, it does not outperform one trained only on native data; however, supplementing native data with synthetic translations does enhance performance.

		\citet{phan-etal-2023-enriching} improve the English-to-Vietnamese \ac{MT} model Mtet by injecting synthetic biomedical parallel text via self-training~\cite{DBLP:journals/corr/abs-1909-13788}, resulting in a system that outperforms strong baselines and enables the creation of ViPubmed and ViMedNLI datasets. Continued pre-training and fine-tuning on these resources lead to ViPubMedT5, which achieves state-of-the-art results in several biomedical \ac{NLP} tasks, further demonstrating the potential of synthetic translation data for advancing low-resource language modeling.

		Finally, \citet{ishigaki-etal-2023-pretraining} pre-train a Japanese BERT model on 2.5M abstracts from Web of Science which were translated into Japanese via Amazon Translate, alongside 1.2M native Japanese abstracts from Wikipedia. Although no comparisons are made with existing \acp{PLM}, experiments involving entity and relation extraction tasks indicate that models that use translated data exhibit superior performance over those trained on native data.

		\section{TransCorpus: A Scalable Translation Framework}

		In this section, we present TransCorpus, a framework designed to facilitate the translation of large-scale corpora into multiple languages. First, the selection of the \ac{MT} toolkit along with its model will be presented, then the model size and context length will be discussed, and finally, the proposed translation workflow will be illustrated.

		\subsection{Machine Translation Framework \& Model Selection}

		To achieve our goal of translating large volumes of text between any two languages, we selected M2M-100 \cite{m2m-100} in conjunction with fairseq as a versatile tool for implementation. Fairseq offers broad support for multilingual tasks and facilitates rapid deployment with multi-GPU processing capabilities. Facebook AI's M2M-100 enables direct translation between languages without using English as an intermediary, making it perfect for converting text from any one of 100 languages to another. Moreover, fairseq's modular framework allows for easy model swapping and the integration of new or specialized translation models. This flexibility customizes the translation process to meet specific needs, such as domain adaptation or the incorporation of additional languages.

		The M2M-100 model is available in three sizes: 418M, 1.2B, and 12B parameters. The smallest model is faster and uses less memory, while the largest requires multiple \acp{GPU} for deployment. Because translation quality improvements come at a quadratic increase in computational cost, we did not consider the 12B model for our experiments. However, since the 418M and 1.2B models differ substantially in translation quality but not as much in computational cost, the next section will compare these two models, focusing primarily on computation time.

		\subsection{Model Size \& Context Length}

		The context length relationship with model complexity is quadratic due to the way attention mechanisms operate in transformers. As context length increases, the number of interactions that the model must account for grows quadratically because every token attends to every other token in the sequence. This means that computational resources, such as time and memory, increase significantly with longer sequences. Overlooking this relationship can lead to inefficient computation times and resource usage, particularly with large datasets or models, resulting in slower processing speeds and potentially prohibitive resource demands. Moreover, \ac{MT} models such as M2M-100 typically trained on sentence pairs might exhibit unexpected behavior if used otherwise. Conversely, having no context would reduce translation to a word-by-word level, resulting in nonsensical outcomes. The following analysis explores document-based and sentence-based translation methods while considering both models sizes on a sample of 1000 life sciences abstracts.


		\begin{figure}[h]
			\def\svgwidth{\linewidth}
			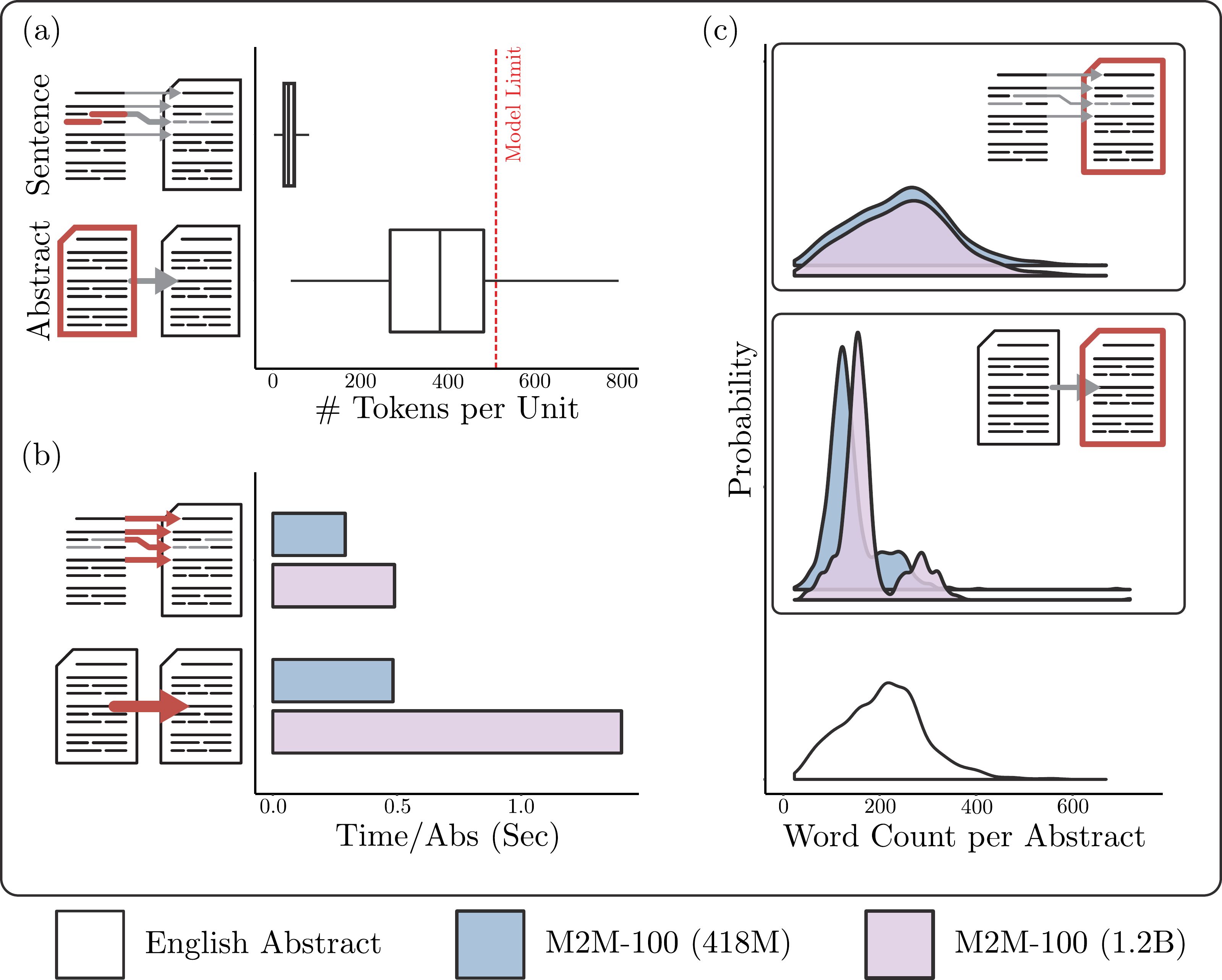
			\caption{\textbf{Translation Method Analysis on a 1000-Abstracts Sample} - (a) Box plot comparing the number of tokens per sentence and abstract, with a red line at 512 tokens representing the maximum token limit that M2M-100 can handle. (b) The average time to translate each abstract using the 418M and 1.2B model versions, comparing sentence-based and document-based translation. (c) Distribution of word count per abstract for both model sizes, displayed with the original English abstract at the bottom when translating by abstract (middle) and by sentence (top). All distributions are normalized to the same scale, so their areas add up to 1.}\label{fig:translation_method}
		\end{figure}

		\hyperref[fig:translation_method]{Figure~\ref*{fig:translation_method}a} clearly demonstrates that when translations are performed by sentence, the distribution tends to favor parallelization because larger differences in sequence length require more padding, leading to wasted computation. \hyperref[fig:translation_method]{Figure~\ref*{fig:translation_method}b} shows that sentence-based translation consistently results in faster processing for any given model size, with the speed advantage becoming more pronounced as the model size increases. It is important to note that the sentence-based approach will tend to scale linearly with the amount of data to be translated, while the document-based approach will highly depend on the document length distribution of a given domain. Finally, the sentence-based approach seems to mimic the original distribution of words per translated document, while the disparity observed in \hyperref[fig:translation_method]{Figure~\ref*{fig:translation_method}c} for the document-based approach was qualitatively reviewed and appears to be partially attributed to a'repetition' problem. \hyperref[app:translation_repeat]{Appendix~\ref*{app:translation_repeat}} shows an observed example. As already mentioned, M2M-100 was trained on sentences pairs, which might explain this behavior.

		While maintaining translation consistency, the sentence-based strategy is adaptable and scalable to various types of documents. Regarding the size of the model, the difference in the translation time in sentence length is negligible compared to the gains reported in quality. This observation led to the decision to adopt the 1.2B model along with a sentence-based translation approach.

		\subsection{Framework Translation Workflow}

\begin{figure}[h]
			\def\svgwidth{\linewidth}
			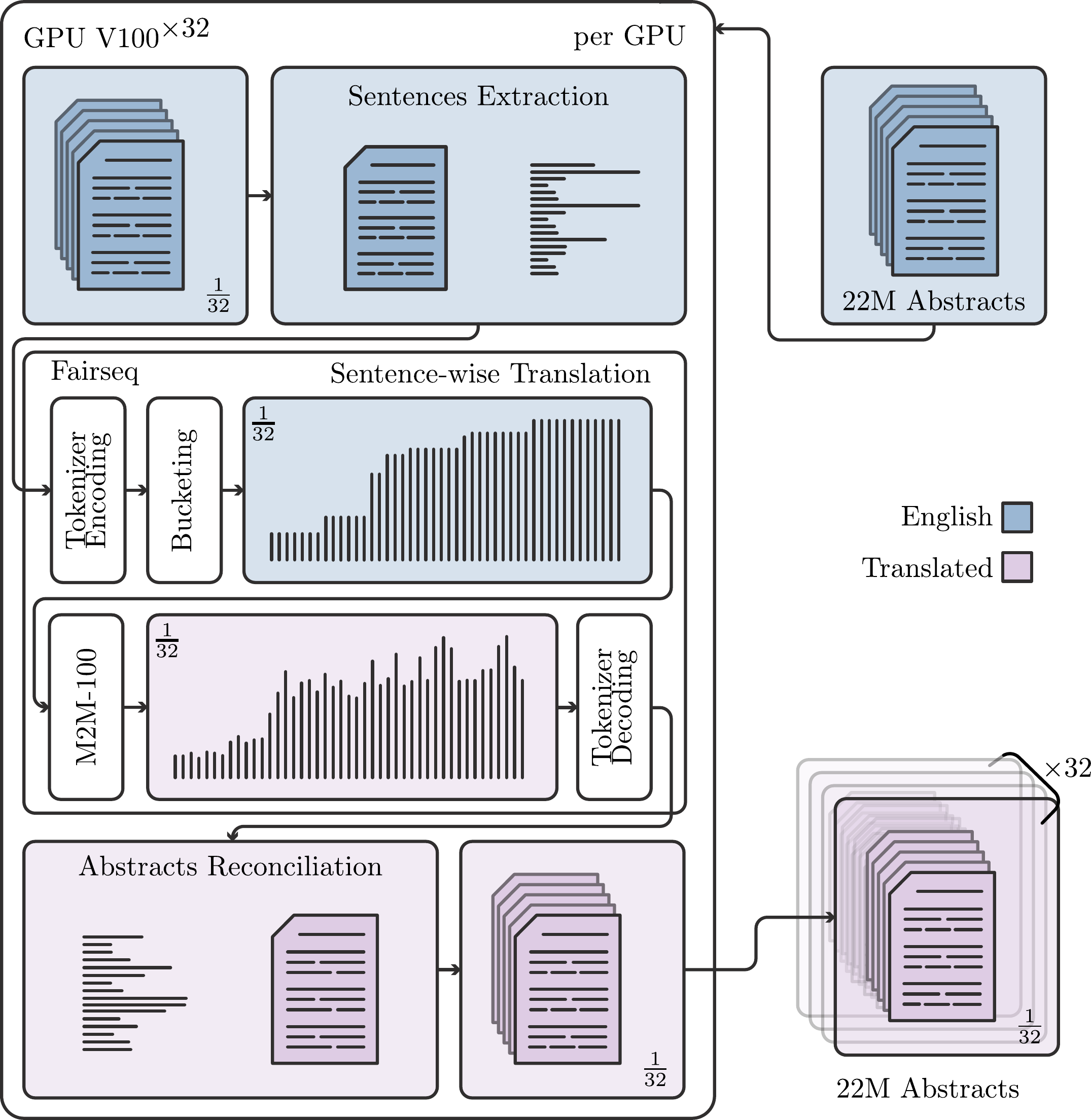
			\caption{\textbf{TransCorpus Translation Workflow} - Illustration of the deployment of TransCorpus on a machine with 32 \acp{GPU}.}
			\label{fig:translation_workflow}
		\end{figure}

		\hyperref[fig:translation_workflow]{Figure~\ref*{fig:translation_workflow}} depicts TransCorpus toolkit applied to an English life sciences corpus consisting of 22M abstracts. First, the corpus is divided and distributed among different machines to parallelize the translation process. Each abstract is then divided into sentences with fairseq handling tokenization as shown in \hyperref[app:sentencesCitation]{Appendix~\ref*{app:sentencesCitation}}. By grouping sentences of the same length, bucketing is employed to minimize padding, thereby avoiding the computational inefficiency that results from juxtaposing long and short sentences. Although it may seem counterintuitive, there is a considerable increase in speed when translating sentences of the same length simultaneously. Once the sentences are translated, they are matched with their respective abstracts and sentence numbers, and the entire corpus is reconciled by concatenating each output of each subprocess. \hyperref[app:transCitation]{Appendix~\ref*{app:transCitation}} shows an abstract translation example. To avoid too short context issues, sentences that contain fewer than 10 characters are concatenated to the following or preceding sentence.

		The TransCorpus toolkit includes a \ac{CLI} that features four primary commands: (1) \code{transcorpus download-corpus [domain]} allows users to fetch a corpus from specified domains, (2) \code{transcorpus preprocess [domain] [target-language] [num-splits]} allows users to optionally preprocess the corpus before translating, and (3) \code{transcorpus translate [domain] [target-language] [num-splits]} handles both preprocessing (if not previously completed) and translating the corpus. Users can perform preprocessing and translation concurrently across numerous \acp{GPU} and processes. A checkpoint recovery feature enables users to resume preprocessing or translation from where it last stopped, which is especially beneficial for academics facing \acp{GPU} usage time limits. Lastly, (4) \code{transcorpus preview [domain] [language1] [language2]} provides a side-by-side document preview of the corpus in two languages. Currently, only the life sciences corpus is ready for download, but by updating the \code{domains.json} file on GitHub with a new corpus URL, additional corpus domains can be quickly integrated into the toolkit. For further details, please check the GitHub repository at \url{https://github.com/jknafou/TransCorpus}.

		\section{TransCorpus-bio-fr: A French Life Sciences Corpus}

		As already highlighted, the limited availability of \ac{DS} \acp{PLM} for certain language/domain pairs presents a notable challenge to the progression of \ac{NLP} tools. Evaluating our framework in a real-world context is complex for several reasons: (1) it is uncommon to find sufficient \ac{DS} for low-resource language/domain pairs that also have datasets for model evaluation, and (2) even when such benchmarks exist, an appropriate \ac{PLM} for comparison might not be available. Fortunately, in the domain of French life sciences, two important papers have recently been published. The first is DrBenchmark~\cite{labrak:hal-04470938} a life sciences benchmark that includes multiple datasets supporting the evaluation of in-domain models. The second is DrBERT~\cite{labrak-etal-2023-drbert}, a French life sciences \ac{PLM}. Comparing our model against DrBERT on life sciences tasks allows us to assess the practical effectiveness of our framework. Indeed, as DrBERT is pre-trained from scratch, it does not rely on an advanced general domain \ac{PLM} such as CamemBERT~\cite{martin-etal-2020-camembert}, which is also the case for most languages.

		\subsection{MEDLINE/PubMed Abstracts Collection}
		For the building of this life sciences corpus, the 2021 \ac{MBR}, encompassing 31M citations, and updates up until April 2021 was downloaded. Then, each citation in the dataset that includes a PMID, a title, and an abstract is kept, subsequently, its raw text is modified by substituting any sequence of one or more whitespace characters with a single space. An example of a title and abstract after modification, as it would appear prior to translation can be found in \hyperref[app:mbrCitation]{Appendix~\ref*{app:mbrCitation}}.

		A considerable amount of citations lacks one of the three essential attributes, i.e. title, abstract, or PMID. Consequently, after filtering the complete dataset, our corpus comprises about 22M abstracts. Despite a few missing unknown values, a comprehensive comparison of our corpus statistics against several models can be found in \hyperref[app:corpusStat]{Appendix~\ref*{app:corpusStat}}. Despite both BioBERT and PubMedBERT \cite{gu2020domain-specific} have a version that also includes \ac{PMC} full-text articles, only those that use PubMed are displayed for a better comparison. This juxtaposition is crucial for understanding the scale of data that similar models have been trained on, which directly impacts their performance and applicability in various \ac{NLP} tasks.

		\subsection{Translated Corpus Statistics}

		The complete translation process was executed using 32 NVIDIA Tesla V100 \acp{GPU} with 32Go of memory each, taking roughly 15 days, which translates to approximately 11,520 \ac{GPU} hours\footnote{These figures are based on an earlier version of TransCorpus; with the latest release of the toolkit, translating the corpus into Spanish requires 7 days using 6 NVIDIA A100 with 80GB of memory each.}. After translation, the resultant raw text file is 36.4GB, containing 221M sentences and 5.25B words. \hyperref[tab:corpus_comparison]{Table~\ref*{tab:corpus_comparison}} compares TransCorpus with the only two French life sciences corpora leveraged for pre-training. The comparison reveals that DrBERT the \ac{SOTA} life sciences \ac{LM} in French, despite it utilizes the largest corpus until now, is about five times smaller than TransCorpus.

		\begin{table}[h]
    \small
    \centering
    \begin{tabularx}{0.48\textwidth}{X|c|c|c}
    \toprule
     & \textbf{\thead{TransCorpus}} & \textbf{\thead{DrBERT\\Corpus}} & \textbf{\thead{CmBERT\\Bio Corpus}}\\
    \midrule
    \textbf{Size}       & 36.4GB & 7.5GB & 2.7GB \\
    \textbf{Sentences}  & 221M   & 54M   & - \\
    \textbf{Words}      & 5.25B   & 1.1B & 413M \\
    \bottomrule
    \end{tabularx}
    \caption{\textbf{Translated Corpus Statistics Compared to French Life Science Corpora} - CmBERT: CamemBERT, '-': Unknown value.}\label{tab:corpus_comparison}
\end{table}

		Even if the corpus size is important, its quality must also be closely monitored. While \ac{MBR} is already considered a benchmark of quality in English as it is used for pre-training models such as BioBERT and PubMedBERT, it is crucial to assess the quality of our translations to make sure that everything has been conducted properly. As depicted in \hyperref[fig:translation_method]{Figure~\ref*{fig:translation_method}c}, a comparable density check of the entire translated corpus reveals a density profile similar to the original corpus. After manually reviewing a randomly chosen set of abstracts, no irregular translation events, such as repetitions, were detected. A few translated abstracts alongside their counterparts originally written in French can be found in \hyperref[app:translation_example]{Appendix~\ref*{app:translation_example}}. The translated corpus is available on Hugging Face at \url{https://huggingface.co/datasets/jknafou/TransCorpus-bio}.

		%

		\section{TransBERT Pre-Training}

		The reasons for pre-training a \ac{LM} from scratch are twofold. Firstly, it enables us to project the usage of our framework on languages that might lack a general domain \ac{PLM}. Secondly, it allows the use of our custom tokenizer, which typically provides enhanced performance for \ac{DS} \acp{LM}.

		\subsection{TransTokenizer Training}\label{sec:tokenizer_training}

		Subword segmentation algorithms aim to split words optimally using probability. Considering the potential addition of more languages in future works, choosing a tokenizer capable of handling specific linguistic features could prove beneficial. In that context, SentencePiece treats whitespaces as regular characters rather than relying on them, which means that it is suited for all kinds of languages. The original SentencePiece implementation\footnote{\url{https://github.com/google/sentencepiece}} \cite{kudo-richardson-2018-sentencepiece} is used to train an Unigram tokenizer with a vocabulary size of 32k and a character coverage set to 0.9995 (default values).

		\subsection{Pre-training Hyperparameters}

		A BERT architecture i.e. a Transformer encoder with 12 hidden layers, each with 12 attention heads of dimension 768, is pre-trained on TransCorpus following RoBERTa \cite{liu-2019-roberta} with an extensive batch size of 8k, an Adam Optimizer \cite{kingma2017adam}, along with 24k warm-up steps and a learning rate of 6e-4. The model was updated for 500k steps on a \ac{MLM} objective function.

		\section{TransBERT-bio-fr: Application to Life Sciences in French}

		This section details the pre-training of TransBERT-bio-fr and compares it with other French \acp{PLM}.

		\subsection{TransTokenizer-bio-fr Training}

		The tokenizer training on TransCorpus-bio-fr took approximately 12 hours on a single machine. As SentencePiece tokenizers require a considerable amount of RAM, a cut-off at 10M translated abstracts were randomly selected in order to train a \ac{DS} tokenizer based on our synthetic translated corpus. An example showcasing the difference between the tokenization of TransTokenizer (ours) and CamemBERT's tokenizer can be found in \hyperref[app:transtokenizer_sample]{Appendix~\ref*{app:transtokenizer_sample}}.

		\subsection{CmTransBERT: Tokenizer Ablation}

		To evaluate the impact of the tokenizer, TransBERT-bio-fr is pre-trained using TransTokenizer-bio-fr while CmTransBERT is pre-trained using CamemBERT's tokenizer. Both models are trained on the TransCorpus-bio-fr, with the same hyperparameters. Prior to fine-tuning our models, the \ac{PPPL} \cite{Salazar_2020} per token and word for each model was computed on a 50 authentic French abstracts. This step confirms the success of the pre-training and provides the go-ahead for the experimental phase. For further details, the results are presented in \hyperref[app:transbert_pppl]{Appendix~\ref*{app:transbert_pppl}}.

		\subsection{Pre-training Statistics}

		Both TransBERT-bio-fr and CmTransBERT were pre-trained for approximately three months using a machine with 3 NVIDIA A100 GPUs, each with 80GB of memory. To ensure a fair pre-training comparison, TransBERT adopted RoBERTa's training methodology, processing 4B sequences over 500k steps with a batch size of 8k. In contrast, DrBERT which represents the best effort for a \ac{LM} for the life sciences domain in French was pre-trained on 310M sequences over 78k steps with a batch size of 4k. Therefore, while TransCorpus-bio-fr's corpus is about five times larger than DrBERT's, TransBERT-bio-fr's overall training data updates are thirteen times larger compared to DrBERT. CamemBERT bio undergoes fewer training updates than DrBERT however, it keeps pre-training CamemBERT, a \ac{PLM} already pre-trained for 100k steps using batch size of 8k. TransBERT-bio-fr and its tokenizer are available on Hugging Face at \url{https://huggingface.co/jknafou/TransBERT-bio-fr}.

		\section{Experimental Setup}

		To compare TransBERT-bio-fr with other \acp{PLM}, we leveraged DrBenchmark which is composed of multiple datasets and tasks. This section describes the experimental setup under which each model was evaluated.

		\subsection{Baseline Models}

		To evaluate our method against strong baselines, we selected the state-of-the-art French PLM, CamemBERT as well as DrBERT the only life sciences PLM pre-trained from scratch in French. As previously noted, CamemBERT bio was excluded from comparison because it is derived from CamemBERT rather than being trained from scratch, whereas our framework is specifically designed for scenarios where \ac{DS} data are available in one language, but resources in the target language are lacking.

		\subsection{DrBenchmark: An Adaptation}

		Common \ac{LM} benchmarks in life sciences are predominantly biomedical or clinical, such as \ac{BLURB} \cite{Gu_2021} and \ac{BLUE} \cite{peng-etal-2019-transfer} in English. In French, only one option was recently published: DrBenchmark. Available in our GitHub, an adaptation of the benchmark containing a few additions such as \ac{HPO} implementation instead of fixed hyperparameters setting, a few data cleaning steps to avoid duplicates, datasets merging to avoid unnecessary small datasets and the implementation of a k-fold cross-validation strategy with multiple iterations to allow for a more robust evaluation. \hyperref[app:drbenchmark_stat-adaptation]{Appendix~\ref*{app:drbenchmark_stat-adaptation}} shows the adapted benchmark datasets statistics, which includes 15 tasks, five of which are classification, six \ac{NER}, two \ac{POS}, and two \ac{STS}.


		\subsection{Statistical Testing}

		Once a metric is computed for each label/class/entity/tag/regression, a statistical test is performed to assess if there is a significant difference between models (1) at the dataset level comparing labels performance across labels and folds and (2) at the task level comparing performances across labels, folds and datasets. For comparisons involving more than two models, the Friedman test is employed, followed by the Nemenyi test. When comparing two models, the Wilcoxon test is used. \hyperref[app:statistical_testing]{Appendix~\ref*{app:statistical_testing}} shows the statistical testing process following \cite{stat_sign_metho} recommended practice for comparing metrics rankings to assess model difference for one or multiple datasets.

		\section{Results \& Discussion}

		\begin{table}[h]
    \setlength{\tabcolsep}{1pt}
    \centering
    \small
    \begin{threeparttable} 
    \begin{tabularx}{.48\textwidth}{cl|>{\centering\arraybackslash}X|>{\centering\arraybackslash}X|>{\centering\arraybackslash}X}
    \toprule
      \multicolumn{2}{c|}{\textbf{Datasets}} & \textbf{CmBERT} & \textbf{DrBERT} & \textbf{TransBERT} \\
    \addlinespace[1pt]
    \toprule
    \multirow{5}{*}{\rotatebox{90}{\textbf{CLS}}} & { DEFT-2020/T2} & \textbf{98.91} (\underline{1}) & 97.55 (\underline{1}) & \underline{98.82} (\textbf{4})  \\
    & { DiaMed} & 64.70 (22)  & \underline{68.89} (\underline{27}) & \textbf{75.32}\tnote{\textbf{*}}(\textbf{55})  \\
    & { FrMedMCQA } & \underline{56.95} (\textbf{14}) & 56.01 (9) & \textbf{57.25} (\underline{10}) \\
    & { MorFITT} & \underline{73.16} (\underline{14}) & 72.74 (8) & \textbf{75.36}\tnote{\textbf{*}}(\textbf{38})  \\
    & { PxCorpus/T2} & \textbf{96.31} (\textbf{11}) & \underline{95.34} (\underline{8}) & \underline{95.34} (7)  \\
    \midrule
    \multirow{6}{*}{\rotatebox{90}{\textbf{NER}}} & { E3C/Clinical} & 74.88 (0) & \underline{75.44} (\underline{1}) & \textbf{76.83} (\textbf{4})  \\
    & { E3C/Temporal} & \underline{85.44} (\textbf{12}) & 83.92 (2) & \textbf{85.73} (\textbf{12})  \\
    & { MantraGSC } & \underline{60.56} (\underline{12}) & 57.80 (8) & \textbf{62.83} (\textbf{16})  \\
    & { PxCorpus/T1} & \underline{92.86} (40) & 92.56 (\underline{66}) & \textbf{95.17}\tnote{\textbf{*}}(\textbf{96})  \\
    & { QUAERO/EMEA } & 84.70 (12) & \underline{84.74} (\underline{13}) & \textbf{85.67}\tnote{\textbf{*}}(\textbf{26})  \\
    & { QUAERO/MdL } & \underline{62.22} (\underline{17}) & 60.71 (5) & \textbf{64.06} (\textbf{29})  \\
    \midrule
    \multirow{2}{*}{\rotatebox{90}{\textbf{POS}}} & { CAS}  & \underline{97.66} (\underline{74}) & 97.56 (50) & \textbf{97.74} (\textbf{75})  \\
    &  { ESSAI } &  \textbf{98.66}\tnote{\textbf{*}}(\textbf{107}) & 98.53 (53) & \underline{98.64} (\underline{71})  \\
    \midrule
    \multirow{2}{*}{\rotatebox{90}{\textbf{STS}}} & { CLISTER} & \textbf{82.80} (\underline{2}) & 75.44 (0) & \underline{82.62} (\textbf{3})  \\
     & { DEFT-2020/T1 } & \textbf{83.95} (\textbf{3}) & 71.69 (0) & \underline{83.46} (\underline{2})  \\
    \bottomrule
    \end{tabularx}
    \begin{tablenotes}
        \item[\textbf{*}] Significant at 0.05 level (Friedman \& Nemenyi test).
    \end{tablenotes}
    \end{threeparttable}
    \caption{\textbf{Performance Evaluation on the French Life Science Datasets} - Table compares the main metrics for each dataset for Classification, Named Entity Recognition, Part-of-Speech Tagging, and Semantic Textual Similarity tasks. F$_1$-score is used for each task as the main metric aside STS which uses R$^2$. In (parentheses) is the count of class/label/entity/tag across all the folds where a model achieved the highest metric. In \textbf{bold} is the highest metric/count while \underline{underlined} text represents the second. CmBERT: CamemBERT.}
    \label{tab:dataset_results}
\end{table}

		\hyperref[tab:dataset_results]{Table~\ref*{tab:dataset_results}} presents models performances across all folds for each dataset with the weighted F$_1$-score for each task except \ac{STS}, which utilizes the R$^2$ metric. Among the 15 datasets evaluated, \ac{TransBERT} outperforms the other models in 10 cases, with statistical significance noted on four occasions. CamemBERT ranks first in five cases, with one statistically significant result. DrBERT fails to achieve the top metric in any dataset and ranks lowest in 11 datasets. In parentheses are the highest labels metric count across all the folds. For instance, in DiaMed, TransBERT secures the highest F$_1$-score for 55 labels over five folds, whereas CamemBERT and DrBERT attain the highest F$_1$-score for 22 and 27 labels, respectively.

		In classification tasks, even though CamemBERT achieves the top performance on two datasets, the differences in metrics and ranking between the models on these datasets are not significant. Conversely, on the DiaMed and MorFITT datasets where TransBERT outperforms, the distinction in metrics and ranking is notable and statistically significant.

		In \ac{NER}, TransBERT leads across all datasets in both metrics and rankings, achieving statistical significance in two instances. In \ac{POS} tasks, the models demonstrate high and closely matched performances, with the lowest-scoring model achieving a weighted F$_1$-score of 97.56. Despite this narrow margin, CamemBERT secures top results for one dataset, showing statistical significance and attaining the highest F$_1$-score across 107 tags in all five folds. In \ac{STS}, CamemBERT and TransBERT perform similarly, with minor differences, obtaining three and two top results, respectively. However, DrBERT performs poorly in this task, particularly with a margin exceeding 10 points in DEFT-2020/T1.

		\subsection{Aggregated Results by Task}

		\begin{table*}[h]
\setlength{\tabcolsep}{1pt}
\centering
\small
\begin{threeparttable} 
\begin{tabularx}{\textwidth}{l|>{\centering\arraybackslash}X>{\centering\arraybackslash}X>{\centering\arraybackslash}X|>{\centering\arraybackslash}X>{\centering\arraybackslash}X>{\centering\arraybackslash}X|>{\centering\arraybackslash}X>{\centering\arraybackslash}X>{\centering\arraybackslash}X}
\toprule
 &\multicolumn{3}{c|}{\textbf{CamemBERT}}&\multicolumn{3}{c|}{\textbf{DrBERT}}&\multicolumn{3}{c}{\textbf{TransBERT}} \\
\cmidrule{2-10}
 & {$\mathbf{P_{w}}$} & {$\mathbf{R_w^{(2)}}$} & {$\mathbf{F_{w}}$} & {$\mathbf{P_{w}}$} & {$\mathbf{R_w^{(2)}}$} & {$\mathbf{F_{w}}$} & {$\mathbf{P_{w}}$} & {$\mathbf{R_w^{(2)}}$} & {$\mathbf{F_{w}}$}  \\
\toprule
\textbf{Classification } & 74.65 & \underline{75.54} & \underline{74.17} & \underline{74.81} & 73.42 & 73.73 & \textbf{75.82}\textsuperscript{**} & \textbf{76.69}\textsuperscript{**} & \textbf{75.71}\textsuperscript{**}  \\
\textbf{\acl{NER} } & \underline{81.23} & \underline{82.13}\textsuperscript{**} & \underline{81.55} & 80.74 & 81.27\textsuperscript{**} & 80.88 & \textbf{83.03}\textsuperscript{**} & \textbf{83.46}\textsuperscript{**} & \textbf{83.15}\textsuperscript{**}  \\
\textbf{\acl{POS} } & \underline{98.31} & \underline{98.29} & \underline{98.29} & 98.20\textsuperscript{**} & 98.18 & 98.18\textsuperscript{**} & \textbf{98.33} & \textbf{98.30} & \textbf{98.31}  \\
\textbf{\acl{STS} } & - & \textbf{83.38} & - & - & 73.56\textsuperscript{**} & - & - & \underline{83.04} & - \\
\bottomrule
\end{tabularx}
\begin{tablenotes}
    \item[\textbf{**}] Significant at 0.01 level (Friedman \& Nemenyi test)
\end{tablenotes}
\end{threeparttable}
\caption{\textbf{Performance Evaluation on the French Life Science by Task} - Weighted Precision, Recall, and F$_1$-scores for each task taking into account each class/label/entity/tag and weighted across all folds and datasets. For \acl{STS}, the weighted $R^{\mathrm{2}}$ is reported. In \textbf{bold} is highest metric/count while \underline{underlined} text represents the second.}
\label{tab:transbert_results_overall}
\end{table*}

		\hyperref[tab:transbert_results_overall]{Table~\ref*{tab:transbert_results_overall}} presents the weighted precision, recall, and F$_1$-score across each task, except for \ac{STS}, which utilizes the R$^2$ metric. TransBERT achieves the best performance for both classification and \ac{NER}, with statistically significant results at the 0.01 level for every metric. CamemBERT secures second place in weighted recall for the \ac{NER} task, also with statistical significance. The difference between CamemBERT and TransBERT in the \ac{POS} task is minimal; though TransBERT leads in terms of the three metrics, the margin between them is slight. In the \ac{STS} task, both CamemBERT and TransBERT do not show statistical significance, while DrBERT comes last with statistical significance.

		With the second more precise classifier, DrBERT ends up having the poorest results in 9 of the 10 metrics. It is worth noting that despite DrBERT is pre-trained on a native French corpus, its sources are quite varied, which could lead to confusion during the pre-training stage for a \ac{LM}. Specifically, it draws from 24 diverse sources such as disease and condition descriptions, clinical cases, meeting reports, health courses, or even optical character recognition data. Beyond this diversity factor, if a provided sequence is too short for the model to deduce a context helping it identify the kind of document it is receiving, this may cause confusion, potentially resulting in ineffective learning. As already mentioned, even if TransBERT corpus is made of synthetic data, it has already been proved that using \ac{MBR} worked in English for pre-training of BioBERT and PubMedBERT.

		\subsection{Tokenizer Ablation Study}

		Although prior work by \citet{labrak-etal-2024-important} explored similar analyses, we identify methodological inconsistencies in their pre-training of 16 \acp{PLM}. Specifically, the use of a fixed time-based stopping criterion resulted in unequal training durations across models, potentially biasing outcomes. Furthermore, the justification for employing reduced batch sizes remains unclear, although computational constraints may have been a contributing factor. To address these limitations, we performed systematic replication under controlled experimental conditions. To our knowledge, this study is the first rigorous examination of how tokenizer impacts \ac{DS} \acp{PLM}, offering insights for optimizing architecture decisions in resource-constrained scenarios.

		\hyperref[tab:ctransbert_results_overall]{Table~\ref*{tab:ctransbert_results_overall}} presents the comprehensive set of weighted main metrics for both models. The results indicate that TransBERT generally outperforms CmTransBERT in almost all tasks, with statistical significance achieved solely in \ac{NER}. This implies that \ac{NER} is more influenced by tokenization compared to other tasks, which seems trivial as \ac{NER} is basically token-based.


		\begin{table*}[h]
	\setlength{\tabcolsep}{2pt}
	\setlength{\tabcolsep}{1pt}
	\centering
	\small
	\begin{threeparttable} 
		\begin{tabularx}{\textwidth}{l|>{\centering\arraybackslash}X>{\centering\arraybackslash}X>{\centering\arraybackslash}X|>{\centering\arraybackslash}X>{\centering\arraybackslash}X>{\centering\arraybackslash}X}
			\toprule
			&\multicolumn{3}{c|}{\textbf{TransBERT}}&\multicolumn{3}{c}{\textbf{CmTransBERT}} \\
			\cmidrule{2-7}
			& {$\mathbf{P_{w}}$} & {$\mathbf{R_w^{(2)}}$} & {$\mathbf{F_{w}}$} & {$\mathbf{P_{w}}$} & {$\mathbf{R_w^{(2)}}$} & {$\mathbf{F_{w}}$}  \\
			\toprule
			\textbf{Classification } & \textbf{75.82} & \textbf{76.69} & \textbf{75.71} & \underline{75.10} & \underline{76.05} & \underline{74.70}  \\
			\textbf{\acl{NER} } & \textbf{83.03}\textsuperscript{**} & \textbf{83.46}\textsuperscript{**} & \textbf{83.15}\textsuperscript{**} & \underline{81.02}\textsuperscript{**} & \underline{82.13}\textsuperscript{**} & \underline{81.44}\textsuperscript{**}  \\
			\textbf{\acl{POS} } & \textbf{98.33} & \textbf{98.30} & \textbf{98.31} & \underline{98.31} & \underline{98.29} & \underline{98.29}  \\
			\textbf{\acl{STS} } & - & \underline{83.04} & - & - & \textbf{84.36} & - \\
			\bottomrule
		\end{tabularx}
		\begin{tablenotes}
		\item[\textbf{**}] Significant at 0.01 level (Wilcoxon test)
		\end{tablenotes}
	\end{threeparttable}
	\caption{\textbf{Ablation study comparing TransBERT and CmTransBERT} - Weighted Precision, Recall, and F$_1$-scores for each task taking into account each class/label/entity/tag and weighted across all folds and datasets. For \ac{STS}, the weighted $R^{\mathrm{2}}$ is reported. In \textbf{bold} is the highest metric/count while \underline{underlined} text represents the second.}
	\label{tab:ctransbert_results_overall}
\end{table*}



		\section{Conclusion \& Contributions}
		This work establishes a rigorous framework for assessing \acp{LM} on \ac{DS} for non-English dataset. It builds upon prior research and extends it to a more comprehensive benchmark that includes a more robust way of evaluating the models by applying \ac{HPO}, multiple training repetition, 5-folds cross-validation, and statistical testing on 15 datasets along with their aggregation by task. It illustrates that employing translated synthetic data for training \ac{DS} \acp{LM} is a viable approach to address the lack of native language data. Our proposed model, TransBERT-bio-fr, outperforms existing \ac{SOTA} models in various life sciences tasks, including classification, \ac{NER}, \ac{POS}, and \ac{STS}.


		In addition to offering a viable methodology to address data scarcity, we release to the public \href{https://github.com/jknafou/TransCorpus}{TransCorpus}, an adaptative toolkit designed to facilitate the translation of large-scale corpora into multiple languages. The resources generated from this work, including \href{https://huggingface.co/datasets/jknafou/TransCorpus-bio}{TransCorpus-bio}, \href{https://huggingface.co/jknafou/TransBERT-bio-fr}{TransBERT-bio-fr} and the code for the pre-training and fine-tuning of the models are also made available on \href{https://github.com/jknafou/TransBERT}{GitHub}.

		\section{Future Work}

		One encouraging direction for future research is to expand our approach to encompass a wider array of languages, especially those that are underrepresented in the life sciences field. Applying our methodology across various linguistic settings will help us better understand its generalizability and any possible constraints. Additionally, creating multilingual models capable of managing several languages within the life sciences sector poses a fascinating challenge. These models might exploit cross-lingual knowledge transfer, allowing for a more efficient use of scarce data resources and promoting a more inclusive global scientific community. Exploring other domains via our toolkit could also yield valuable insights into the adaptability of our approach.

		Another path for future research is an extensive comparison between our method and the latest generative \acp{LLM} on identical datasets. Such a comparison would yield valuable understanding of the trade-offs between specialized, domain-focused models and more general, resource-heavy models \acp{LLM}. Assessing performance, efficiency, and cost-effectiveness across different life sciences tasks would help researchers and practitioners in making informed decisions. Furthermore, this analysis could highlight the possibility of integrating the strengths of both approaches.

		A promising direction for upcoming research involves exploring the use of generative \acp{LLM} to create synthetic data for training \ac{DS} models, as an alternative to our translation-based method. This approach could yield more varied and nuanced datasets, encapsulating intricate \ac{DS} knowledge and linguistic patterns. Assessing the quality, reliability, and possible biases of \acp{LM}-generated synthetic data in comparison to translated data could offer valuable insights into data augmentation strategies for low-resource domains and languages.

		\section*{Limitations}
		\subsection{Baseline Model}
		While TransBERT-bio-fr got better results than CamemBERT, it would be interesting to see if DrBERT, the \ac{DS} baseline \ac{PLM} would have had better results if it had undergone a proper pre-training process (e.g., 500k steps, 8k batch size, etc.). Also, in order to extend our approach to languages where high quality \acp{PLM} are available, it would have been interesting to compare a pre-training continuation of CamemBERT on TransCorpus-bio-fr with CamemBERT bio.

		\subsection{In-Domain/Language Generalization}
		Even though our benchmark includes a broad range of datasets and tasks, it is impossible to cover every potential application or future development in the field. The performance of our model, while impressive within the scope of our study, may not necessarily be consistent across all possible tasks or datasets in the life sciences domain. Additionally, the idea of a universally 'best' model is inherently flawed in the realm of \ac{NLP}. Different models might excel in particular contexts or specific types of tasks, and their performance can be affected by factors such as domain specificity, data distribution, and the nuances of individual use cases. What works optimally in one scenario may not be the best choice in another, emphasizing the need for context-specific model evaluation and selection. It is also important to recognize that the fast-paced advancements in \ac{NLP} research could lead to new architectures, pre-training techniques, or fine-tuning strategies that may surpass our current model in certain aspects. The dynamic nature of the field requires ongoing evaluation and comparison against new innovations.

		\subsection{Other Domains Generalization}
		Although our model, which was trained on translated synthetic data within the life sciences corpus, shows encouraging generalization towards other domains, it is important to recognize the constraints when extrapolating these results to other areas. The success of our method in addressing the lack of native language data in life sciences should not be automatically expected to apply to other specialized sectors such as finance, law, or engineering. Each field presents its own unique linguistic hurdles, specialized terminologies, and \ac{DS} conceptual frameworks that general-purpose \ac{MT} systems might not handle effectively. The quality and relevance of translated synthetic data can differ greatly between domains, possibly affecting the model's performance. Moreover, the subtleties of \ac{DS} language use, such as idiomatic phrases, technical lingo, and context-dependent meanings, may not be accurately preserved in translated data, which could lead to misunderstandings or errors in other fields. Additionally, the success of our approach may depend on the degree to which translatable concepts are within a given domain, which can vary greatly. For example, concepts that are highly specific to a culture or legally bound in sectors like law or social sciences might pose particular difficulties for this approach. Hence, even if our results suggest a promising avenue for mitigating language resource shortages in specialized fields, further research is essential to confirm the broad applicability of this method across various domains, each with its own distinct linguistic and conceptual challenges.

		\subsection{Other Languages Generalization}
		While our study highlights the effectiveness of employing synthetic translated data for training \acp{LM} in the field of life sciences in French, caution is warranted when applying these findings to other languages, especially those with limited resources. We believe that the success of our method is highly dependent on the quality and availability of \ac{MT} systems for the target language, which can differ greatly among various language pairs. Even if M2M-100 has a great potential to secure relatively great results in low-resource languages compared to other models, some language pairs often lack strong machine translation models, which can undermine the quality of the translated synthetic data. For instance, \citet{10.1162/tacl_a_00615} show that \ac{MT} models can encounter difficulties with hallucinations, especially in low-resource language directions and when translating out of English languages, which may result in misleading outcomes. Additionally, the linguistic gap between the source language and the target language can greatly affect the effectiveness of the approach. Languages with different syntactic frameworks, morphological structures, or writing systems might pose additional difficulties in maintaining semantic subtleties and \ac{DS} language during translation. Furthermore, the cultural and scientific context embedded in the original material might not always have direct counterparts in the target language or culture, which could result in meaning loss or the introduction of biases. Although our findings indicate a potential solution for addressing the deficit of scientific corpora in some languages, the method's suitability across different linguistic contexts requires thorough evaluation and additional investigation.

		\section*{Acknowledgments}
		The work presented in this paper was performed under the umbrella of the \href{https://www.hes-so.ch/swiss-ai-center}{Swiss AI Center} of the HES-SO, thanks to the partial support of the METAPLANTCODE project (SNF Biodiversa+ \#216811, 2024-2027) and the \href{https://elixir-europe.org/platforms/data}{ELIXIR Data Platform}. The experiments were computed on Baobab, Geneva's High Performance Computing infrastructure for academic research.

		\bibliography{anthology,custom,references-thesis}

		\clearpage
		\appendix
		\begin{figure*}[h]
  \section{Appendix}
  \subsection{Example of an English Abstract}\label{app:mbrCitation}
  \begin{tcolorbox}[colback=gray!20,colframe=gray!80]
  \textbf{PMID:} 44 \vspace{8pt} \\
  \textbf{Title:} The origin of the alkaline inactivation of pepsinogen. \vspace{8pt} \\
  \textbf{Abstract:} Above pH 8.5, pepsinogen is converted into a form which cannot be activated to pepsin on exposure to low pH. Intermediate exposure to neutral pH, however, returns the protein to a form which can be activated. Evidence is presented for a reversible, small conformational change in the molecule, distinct from the unfolding of the protein. At the same time, the molecule is converted to a form of limited solubility, which is precipitated at low pH, where activation is normally seen. The results are interpreted in terms of the peculiar structure of the pepsinogen molecule. Titration of the basic NH2-terminal region produced an open form, which can return to the native form at neutral pH, but which is maintained at low pH by neutralization of carboxylate groups in the pepsin portion.
  \end{tcolorbox}
  \caption[Example of a Citation From the MBR Database]{\textbf{Example of a Citation From the MBR Database}}\label{fig:mbrCitation}
\end{figure*}

		\begin{table*}[h]
\subsection{Corpus Statistics}\label{app:corpusStat}
\centering
\begin{threeparttable} 
\begin{tabularx}{\textwidth}{l|>{\centering\arraybackslash}X|>{\centering\arraybackslash}X|>{\centering\arraybackslash}X|>{\centering\arraybackslash}X}
\toprule
 & \textbf{\thead{Corpus Before\\Translation}} & \textbf{\thead{BERT}} & \textbf{\thead{BioBERT}} & \textbf{\thead{PubMedBERT}} \\ 
\midrule
\textbf{Abstracts}                 & 22M    & N/A   & -     & 14M   \\
\textbf{Size}                      & 30.2GB & 16GB  & -     & 21GB  \\
\textbf{Sentences}                 & 202M   & -     & -     & -     \\
\textbf{Words}                     & 4.4B   & 3.3B  & 4.5B  & 3.1B  \\
\textbf{Tokens}                    & 6.7B   & -     & -     & -    \\ 
\bottomrule
\end{tabularx}
\begin{tablenotes}
    \item \textbf{-}: Unknown value
    \item \textbf{N/A}: Not Applicable
\end{tablenotes}
\end{threeparttable}
\caption{\textbf{Statistics of English Life Science Corpora Used to Pre-Train Different Models} - Tokens number is computed using a \ac{BERT} cased tokenizer.}\label{tab:corpusStat}
\end{table*}

		\begin{figure*}[h]
\subsection{Example of a Translation with Repetition}\label{app:translation_repeat}
\begin{tcolorbox}[colback=gray!20,colframe=gray!80]
    \textbf{Model Size:} 418M \vspace{8pt} \\
    \textbf{Translation Approach:} By abstract\vspace{8pt} \\
    \textbf{Abstract:}
    Des modifications structurelles et fonctionnelles dans les ovaries de l'ovaire de contrôle des ovaries des ovaries de contrôle des ovaries des ovaries des ovaries des ovaries des ovaries des ovaries des ovaries des ovaries des ovaries des ovaries des ovaries des ovaries des ovaries des ovaries des ovaries des ovaries des ovaries des ovaries des ovaries des ovaries des ovaries des ovaries des ovaries des ovaries des ovaries des ovaries des ovaries des ovaries des ovaries des ovaries des ovaries des ovaries des ovaries des ovaries des ovaries des ovaries des ovaries des ovaries des ovaries des ovaries des ovaries des ovaries des ovaries des ovaries des ovaries des ovaries des ovaries des ovaries des ovaries des ovaries des ovaries des ovaries des ovaries des ovaries des ovaries des ovaries des ovaries des ovaries des ovaries des ovaries des ovaries des ovaries des ovaries des ovaries.
    \end{tcolorbox}
    \caption[Example of a Translation: 418M, By Abstract (With Repetition)]{\textbf{Example of a Translation: 418M, By Abstract (With Repetition)}}\label{fig:translation_repeat}
\end{figure*}

		\begin{figure*}[h]
\subsection{Example of a Tokenized Abstract}\label{app:sentencesCitation}
\small
\begin{tcolorbox}[colback=gray!20,colframe=gray!80]
\textbf{PMID:} 44 \vspace{8pt} \\
\textbf{Sentence 1:} The origin of the alkaline inactivation of pepsinogen.\par \textcolor{bulgarianrose}{\small ['\_The', '\_origin', '\_of', '\_the', '\_alkal', 'ine', '\_in', 'activ', 'ation', '\_of', '\_pep', 'sin', 'ogen', '.']}\vspace{8pt} \\
\textbf{Sentence 2:} Above pH 8.5, pepsinogen is converted into a form which cannot be activated to pepsin on exposure to low pH.\par \textcolor{bulgarianrose}{\small ['\_Ab', 'ove', '\_pH', '\_8.', '5,', '\_pep', 'sin', 'ogen', '\_is', '\_convert', 'ed', '\_into', '\_a', '\_form', '\_which', '\_cannot', '\_be', '\_activ', 'ated', '\_to', '\_pep', 'sin', '\_on', '\_expos', 'ure', '\_to', '\_low', '\_pH', '.']}\vspace{8pt} \\
\textbf{Sentence 3:} Intermediate exposure to neutral pH, however, returns the protein to a form which can be activated. \par \textcolor{bulgarianrose}{\small ['\_Inter', 'medi', 'ate', '\_expos', 'ure', '\_to', '\_neutral', '\_pH', ',', '\_however', ',', '\_retur', 'ns', '\_the', '\_protein', '\_to', '\_a', '\_form', '\_which', '\_can', '\_be', '\_activ', 'ated', '.']}\vspace{8pt} \\
\textbf{Sentence 4:} Evidence is presented for a reversible, small conformational change in the molecule, distinct from the unfolding of the protein. \par \textcolor{bulgarianrose}{\small ['\_Ev', 'idence', '\_is', '\_present', 'ed', '\_for', '\_a', '\_re', 'vers', 'ible', ',', '\_small', '\_conform', 'ational', '\_change', '\_in', '\_the', '\_mol', 'ec', 'ule', ',', '\_distin', 'ct', '\_from', '\_the', '\_un', 'fold', 'ing', '\_of', '\_the', '\_protein', '.']}\vspace{8pt} \\
\textbf{Sentence 5:} At the same time, the molecule is converted to a form of limited solubility, which is precipitated at low pH, where activation is normally seen.\par \textcolor{bulgarianrose}{\small ['\_At', '\_the', '\_same', '\_time', ',', '\_the', '\_mol', 'ec', 'ule', '\_is', '\_convert', 'ed', '\_to', '\_a', '\_form', '\_of', '\_limited', '\_sol', 'ub', 'ility', ',', '\_which', '\_is', '\_precip', 'itat', 'ed', '\_at', '\_low', '\_pH', ',', '\_where', '\_activ', 'ation', '\_is', '\_norm', 'ally', '\_seen', '.']}\vspace{8pt} \\
\textbf{Sentence 6:} The results are interpreted in terms of the peculiar structure of the pepsinogen molecule.\par \textcolor{bulgarianrose}{\small ['\_The', '\_results', '\_are', '\_interpret', 'ed', '\_in', '\_terms', '\_of', '\_the', '\_pec', 'uliar', '\_structure', '\_of', '\_the', '\_pep', 'sin', 'ogen', '\_mol', 'ec', 'ule', '.']}\vspace{8pt} \\
\textbf{Sentence 7:} Titration of the basic NH2-terminal region produced an open form, which can return to the native form at neutral pH, but which is maintained at low pH by neutralization of carboxylate groups in the pepsin portion.\par \textcolor{bulgarianrose}{\small ['\_T', 'itr', 'ation', '\_of', '\_the', '\_basic', '\_NH', '2-', 'termin', 'al', '\_region', '\_produc', 'ed', '\_an', '\_open', '\_form', ',', '\_which', '\_can', '\_return', '\_to', '\_the', '\_n', 'ative', '\_form', '\_at', '\_neutral', '\_pH', ',', '\_but', '\_which', '\_is', '\_mainta', 'ined', '\_at', '\_low', '\_pH', '\_by', '\_neutr', 'aliz', 'ation', '\_of', '\_car', 'box', 'yl', 'ate', '\_groups', '\_in', '\_the', '\_pep', 'sin', '\_por', 'tion', '.']}
\end{tcolorbox}
\caption[Example of Sentence \& Word Tokenization]{\textbf{Example of Sentence \& Word Tokenization}}\label{fig:sentencesCitation}
\end{figure*}

		\begin{figure*}[h]
\subsection{Example of a Translated Citation}\label{app:transCitation}
\begin{tcolorbox}[colback=gray!20,colframe=gray!80]
\textbf{PMID:} 44 \vspace{8pt} \\
\textbf{Title:} L'origine de l'inactivation alcaline du pepsinogène. \vspace{8pt} \\
\textbf{Abstract:} Au-dessus du pH de 8,5, le pepsinogène est converti en une forme qui ne peut pas être activée en pepsine en cas d'exposition à un pH bas. L'exposition intermédiaire au pH neutre, cependant, renvoie la protéine à une forme qui peut être activée. Des preuves sont présentées pour un changement réversible, de petite conformation dans la molécule, distinct du déploiement de la protéine. Dans le même temps, la molécule est convertie en une forme de solubilité limitée, qui est précipitée à faible pH, où l'activation est normalement observée. Les résultats sont interprétés en termes de la structure particulière de la molécule de pepsinogène. La titration de la région terminale de base NH2 produit une forme ouverte, qui peut revenir à la forme native à pH neutre, mais qui est maintenue à un pH bas par la neutralisation des groupes carboxylés dans la portion de pepsine.
\end{tcolorbox}
\caption[Example of Title and Abstract Citation From the MBR Database Translated in French]{\textbf{Example of Title and Abstract Citation From the MBR Database Translated in French \cite{PMID:44}}}\label{fig:transCitation}
\end{figure*}

\onecolumn
\begin{longtable}{p{\textwidth}}
\subsection{Translation Examples Compared to True French Abstracts}\label{app:translation_example}
\small
\begin{tcolorbox}[colback=gray!20,colframe=gray!80]
\textbf{Original (PMID:33739270)}\\
Le foie assure une grande partie du métabolisme des xénobiotiques. Ses particularités en font pourtant une cible privilégiée pour des composés toxiques. Les hépatotoxicités des xénobiotiques, ces molécules étrangères à notre organisme, constituent un vrai défi pour les cliniciens, l’industrie pharmaceutique, et les agences de santé. à la différence des hépatotoxicités intrinsèques, prévisibles et reproductibles, les hépatotoxicités idiosyncrasiques surviennent de manière non prévisible. La physiopathologie des hépatotoxicités idiosyncrasiques à médiation immune reste la moins bien connue. Le développement d’outils qui permettent désormais d’améliorer la prédiction et la compréhension de ces atteintes hépatiques paraît être une approche prometteuse pour identifier des facteurs de risque, et de nouveaux mécanismes de toxicité.
\end{tcolorbox}\\
\small
\begin{tcolorbox}[colback=gray!20,colframe=gray!80]
\textbf{Translated (PMID:33739270)}\\
Le foie assure une grande partie du métabolisme des xénobiotiques grâce à son équipement enzymatique considérable, à sa localisation anatomique et à sa vascularisation abondante. Cependant, ces différentes caractéristiques en font également une cible privilégiée pour les composés toxiques, en particulier dans le cas d'un métabolisme toxique. L'hépatotoxicité induite par les xénobiotiques est une cause majeure de lésions hépatiques et un véritable défi pour les cliniciens, l'industrie pharmaceutique et les agences de santé. Les hépatotoxicités intrinsèques, c'est-à-dire les hépatotoxicités prévisibles et reproductibles qui se produisent à des doses limites, sont distinguées des hépatotoxicités idiosyncratiques, qui se produisent de manière imprévisible chez les personnes présentant des sensibilités individuelles. Parmi eux, la pathophysiologie de l'hépatotoxicité immunomédiée idiosyncratique n'est toujours pas claire. Cependant, le développement d’outils visant à améliorer la prévision et la compréhension de ces troubles peut ouvrir des voies pour l’identification de facteurs de risque et de nouveaux mécanismes de toxicité.
\end{tcolorbox}\\
\small
\begin{tcolorbox}[colback=gray!20,colframe=gray!80]
\textbf{Original (PMID:32334967)}\\
La tuberculose est due au complexe M. tuberculosis, dont la croissance lente entraîne un long délai de rendu des tests phénotypiques utilisés pour le diagnostic bactériologique. La biologie moléculaire a réduit considérablement ce délai, notamment grâce au déploiement de la méthode Xpert® MTB/RIF (Cepheid) qui permet de détecter le complexe M. tuberculosis et la résistance à la rifampicine en 2 heures. D’autres tests détectant en plus la résistance à l’isoniazide et aux antituberculeux de seconde ligne ont été développés. Cependant, les performances de ces tests sont nettement moins bonnes si l’examen microscopique est négatif. Il est donc crucial de restreindre leur indication aux fortes suspicions cliniques. Les tests de détection de la résistance n’explorent que certaines positions caractérisées ; or, toutes les mutations responsables de l’acquisition de résistance ne sont pas connues. De plus, les performances sont variables pour les différents antituberculeux. L’avènement du séquençage génomique est une perspective prometteuse. La faisabilité en routine doit encore être évaluée et l’analyse des données reste à standardiser. L’essor des techniques de biologie moléculaire a révolutionné le diagnostic de la tuberculose et de la résistance. Cependant, elles restent des tests de dépistage dont les résultats doivent être confrontés aux méthodes phénotypiques de référence.
\end{tcolorbox}\\
\small
\begin{tcolorbox}[colback=gray!20,colframe=gray!80]
\textbf{Translated (PMID:32334967)}\\
La tuberculose est causée par le complexe M. tuberculosis. Sa croissance lente retarde le diagnostic bactériologique basé sur des tests phénotypiques. La biologie moléculaire a considérablement réduit ce retard, notamment grâce au déploiement du test Xpert® MTB/RIF (Cepheid), qui détecte le complexe de M. tuberculose et la résistance à la rifampicine en 2 heures. D'autres tests détectant la résistance à l'isoniazide et aux médicaments antituberculeux de deuxième ligne ont été développés. Cependant, les performances des tests moléculaires sont considérablement réduites si le dépistage de la microscopie de bacille acide rapide est négatif. Il est donc crucial de limiter leur indication à de fortes suspicions cliniques. Les tests de détection de la résistance n'explorent que certaines positions caractérisées; cependant, toutes les mutations de résistance aux médicaments ne sont pas connues. En outre, les performances varient pour différents médicaments antituberculeux. L’avènement de la séquençage génomique est prometteur. Son intégration dans le flux de travail de routine doit encore être évaluée et l’analyse des données doit encore être normalisée. La montée des techniques de biologie moléculaire a révolutionné le diagnostic de la tuberculose et de la résistance aux médicaments. Cependant, ils restent des tests de dépistage; les résultats doivent encore être confirmés par des méthodes de référence phénotypiques.
\end{tcolorbox}\\
\small
\begin{tcolorbox}[colback=gray!20,colframe=gray!80]
\textbf{Original (PMID: 33742585)}\\
Dans un souci d’amélioration de la qualité de vie des personnes atteintes de maladie chronique, les pratiques de soins se sont enrichies de l’éducation thérapeutique du patient (ETP). Celle-ci vise l’acquisition de savoirs et de compétences plurielles par les malades pour favoriser une gestion optimale de la pathologie au quotidien et des changements qui en découlent, en limitant les répercussions négatives sur leur autonomie et leur bien-être. Le sujet est placé au cœur de son dispositif, en position de décision et de responsabilité, et collabore activement avec les différents acteurs de soins. L’ETP implique donc la prise en compte de la dimension psychique du patient, en s’appuyant sur la psychologie et des concepts fondamentaux pour sa mise en œuvre.
\end{tcolorbox}\\
\small
\begin{tcolorbox}[colback=gray!20,colframe=gray!80]
\textbf{Translated (PMID: 33742585)}\\
Dans un effort pour améliorer la qualité de vie des personnes atteintes de maladies chroniques, les pratiques de soins ont été enrichis par l'éducation thérapeutique des patients (TPE). Cela vise à l’acquisition de connaissances et de compétences plurielles par les patients, ce qui favorise une gestion optimale de la maladie sur une base quotidienne et des changements qui en découlent, en limitant leurs répercussions négatives sur leur autonomie et leur bien-être. Le sujet est placé au cœur du système, dans une position de décision et de responsabilité, et collabore activement avec les différents acteurs de la santé. Le TPE implique donc la prise en compte de la dimension psychologique du patient, en utilisant la psychologie et les concepts fondamentaux pour sa mise en œuvre.
\end{tcolorbox}\\
\end{longtable}

		\begin{figure*}[h]
\subsection{Tokenizers Comparison Example}\label{app:transtokenizer_sample}
\begin{tcolorbox}[colback=gray!20,colframe=gray!80]
\textbf{Entity}: ['infarctus', 'du', 'myocarde,'] (3 words)\\
\textbf{TransTokenizer}: ['▁infarctus', '▁du', '▁myocarde', ','] (4 tokens)\\
\textbf{CamemBERT}: ['▁inf', 'arc', 'tu', 's', '▁du', '▁my', 'oc', 'arde', ','] ($\Delta$+5)
\end{tcolorbox}
\caption[CamemBERT Vs TransTokenizer Sample]{\textbf{CamemBERT Vs TransTokenizer Sample} - An example of tokenization shows that the tokenizer of TransBERT (i.e., TransTokenizer) requires less tokens than the tokenizer of CamemBERT to encode the same sequence.}\label{fig:transtokenizer_sample}
\end{figure*}

		\begin{table*}[h]
\subsection{Pseudo-Perplexity Comparison Across Models}\label{app:transbert_pppl}
\centering
\begin{tabularx}{\textwidth}{l|>{\centering\arraybackslash}X|>{\centering\arraybackslash}X|>{\centering\arraybackslash}X|>{\centering\arraybackslash}X}
\toprule
 & \textbf{TransBERT} & \textbf{CmTransBERT} & \textbf{CamemBERT} & \textbf{DrBERT} \\
\midrule
PPPL$_{token}$     & 6.00  & 4.14 & \textbf{174.42}  & 8.30 \\
PPPL$_{word}$        & 11.71 & 8.59 & \textbf{2474.88} & 17.55 \\
\cmidrule{2-5}
$n_{sentence}$     &   \multicolumn{4}{c}{376} \\
\cmidrule{2-5}
$n_{word}$          &   \multicolumn{4}{c}{\num{9204}}\\
\cmidrule{2-5}
$n_{token}$        & \num{12640} & \textbf{\num{13934}} & \textbf{\num{13934}} & \num{12459} \\
\bottomrule
\end{tabularx}
\caption[Pseudo-Perplexity Comparison Across Models]{\textbf{Pseudo-Perplexity Comparison Across Models} - Pseudo-Perplexity across models, with the highest uncertainty highlighted in bold.}\label{tab:transbert_pppl}
\end{table*}

		\begin{table*}[h]
\subsection{Downstream Tasks Summary}\label{app:drbenchmark_stat-adaptation}
\centering
\begin{tabularx}{\textwidth}{l|>{\centering\arraybackslash}X|>{\centering\arraybackslash}X|>{\centering\arraybackslash}X|l}
\toprule
\textbf{Name} & \textbf{Task} & \textbf{Instance} & \textbf{Label} & \textbf{Source} \\
\midrule
CAS & POS & \num{86805} & 30T & CC \\
\midrule
CLISTER & STS & \num{1000} & 0 to 5 & CC \\
\midrule
\multirow{2}{*}{DEFT-2020} & STS & \num{1009} & 0 to 5 & \multirow{2}{*}{\begin{tabular}[c]{@{}l@{}}CC, encyclopedia \&\\ drug\end{tabular}} \\
\cmidrule{2-4}
 & CLS & \num{1100} & 3C & \\
\midrule
DiaMed & CLS & \num{726} & 15C & CC \\
\midrule
E3C/Clinical & \multirow{2}{*}{NER} & \num{3270} & 1E & \multirow{2}{*}{CC} \\ 
\cmidrule{1-1}\cmidrule{3-4}
E3C/Temporal &  & \num{5756} & 5E & \\
\midrule
\multirow{2}{*}{ESSAI} & \multirow{2}{*}{POS} & \multirow{2}{*}{\num{150269}} & \multirow{2}{*}{29T} & Clinical Trial Protocols \\
&&&&\\
\midrule
FrenchMedMCQA & CLS & \num{3102} & 5C & Pharmacy Exam \\
\midrule
\multirow{2}{*}{MantraGSC} & \multirow{2}{*}{NER} & \multirow{2}{*}{\num{879}} & \multirow{2}{*}{7E} & Biomedical, Drug \& Patent \\
&&&&\\
\midrule
MorFITT & CLS & \num{5115} & 12L & Biomedical \\
\midrule
\multirow{2}{*}{PxCorpus} & NER & \num{11465} & 30E & \multirow{2}{*}{Drug} \\
\cmidrule{2-4}
 & CLS & \num{1727} & 4C & \\
\midrule
QUAERO/EMEA & \multirow{2}{*}{NER} & \num{6001} & \multirow{2}{*}{10E} & \multirow{2}{*}{Drug \& Biomedical} \\
\cmidrule{1-1}\cmidrule{3-3}
QUAERO/Medline & & \num{6765} &  &  \\

\bottomrule
\end{tabularx}
\caption{\textbf{DrBenchmark Adaptation: Data \& Tasks Summary} - By alphabetical order - Overall, every model tested will be evaluated using cross-validation on 15 distinct datasets covering a broad range of tasks. In the Label column, \code{C} indicates a class within a multi-class framework, while \code{L} denotes the count of potential labels in a multi-label classification, \code{T} tag and \code{E} entity. The instance count reflects the number of positive \code{C}, \code{L}, \code{T} or \code{E}. In the source column \code{CC} stands for Clinical Cases.}
\label{tab:drbenchmark_stat-adaptation}
\end{table*}

		\subsection{Statistical Testing}\label{app:statistical_testing}
\begin{figure*}[h]
\def\svgwidth{\linewidth}
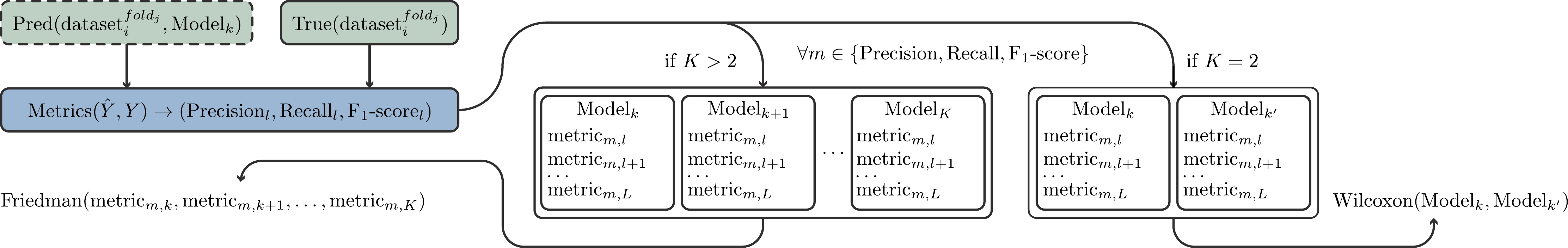\hfill
\caption{\textbf{Statistical Testing} - In order to compare more than two models, the Friedman test is used to determine if there is a significant difference between models, if so, the Nemenyi post-hoc test is used to determine which models are significantly different. For two models, the Wilcoxon test is used.}
\label{fig:statistical_testing}
\end{figure*}

	\end{document}